# Evaluation of an AI system for the automated detection of glaucoma from stereoscopic optic disc photographs: the European Optic Disc Assessment Study

Thomas W. Rogers[1,*], Nicolas Jaccard[1,*], Francis Carbonaro[2], Hans G. Lemij[3], Koenraad A. Vermeer[3], Nicolaas J. Reus[3,4], Sameer Trikha[1,5]

[1] Visulytix Ltd, Screenworks, Highbury Grove, Highbury East, London, N5 2EF, United Kingdom.

[2] Mater Dei Hospital, Triq Dun Karm, L'Imsida, Malta.

[3] Rotterdam Eye Hospital, Rotterdam, The Netherlands.

[4] Amphia Hospital, Department of Ophthalmology, Breda, The Netherlands.

[5] King's College Hospital NHS Foundation Trust, London, SE5 9RS, United Kingdom.

[*] These authors contributed equally to this work

**Support**

All equipment and funding for this work was provided by Visulytix Ltd.

# Abstract

**Objectives:** To evaluate the performance of a deep learning based Artificial Intelligence (AI) software for detection of glaucoma from stereoscopic optic disc photographs, and to compare this performance to the performance of a large cohort of ophthalmologists and optometrists.

**Methods:** A retrospective study evaluating the diagnostic performance of an AI software (Pegasus v1.0, Visulytix Ltd., London UK) and comparing it to that of 243 European ophthalmologists and 208 British optometrists, as determined in previous studies, for the detection of glaucomatous optic neuropathy from 94 scanned stereoscopic photographic slides scanned into digital format.

**Results:** Pegasus was able to detect glaucomatous optic neuropathy with an accuracy of 83.4% (95% CI: 77.5-89.2). This is comparable to an average ophthalmologist accuracy of 80.5% (95% CI: 67.2-93.8) and average optometrist accuracy of 80% (95% CI: 67-88) on the same images. In addition, the AI system had an intra-observer agreement (Cohen's Kappa, κ) of 0.74 (95% CI: 0.63-0.85), compared to 0.70 (range: -0.13-1.00; 95% CI: 0.67-0.73) and 0.71 (range: 0.08-1.00) for ophthalmologists and optometrists, respectively. There was no statistically significant difference between the performance of the deep learning system and ophthalmologists or optometrists. There was no statistically significant difference between the performance of the deep learning system and ophthalmologists or optometrists.

**Conclusion:** The AI system obtained a diagnostic performance and repeatability comparable to that of the ophthalmologists and optometrists. We conclude that deep learning based AI systems, such as Pegasus, demonstrate significant promise in the assisted detection of glaucomatous optic neuropathy.

# Introduction

Primary open-angle glaucoma is a major cause of irreversible blindness, affecting an estimated 80 million people globally by the year 2020.[1] A large proportion of these patients are thought to reside in lower income countries with limited access to care.[1] It is commonly accepted that the early recognition of glaucoma, such as through the detection of early structural changes of the optic nerve from fundus images, or by the detection of raised intraocular pressure, can lead to prompt treatment and the reduction in the burden of sight loss. Whilst primary eye care has taken an increased role in the detection of glaucoma, a number of studies have shown high levels of inter-observer variability in the assessment of optic nerve pathology.[2-4]

The rising prevalence of glaucoma, in part due to ageing populations, imparts a significant burden on already stretched secondary eye care clinics globally. It is estimated that the National Health Service (NHS) in the United Kingdom will be required to make £20 billion efficiency savings in the coming years.[5,6] Traditional models of care, whereby primary eye care providers simply refer patients to specialists, have shown mixed results. In 2010, Lockwood et al. showed a positive predictive rate of 0.37 in the diagnosis of glaucoma or ocular hypertension in patients referred directly from community optometrists.[7] One potential consequence of this is increase in referrals to Hospital Eye Services (HES).

In more recent times, direct referrals from optometrists to HES have been replaced by the establishment of glaucoma referral refinement schemes as a gold standard to which eye care organizations should abide.[8] A number of such schemes exist across the United Kingdom, notably in Gloucestershire, Nottingham and Portsmouth. The Portsmouth based scheme has been particularly effective in streamlining referrals and increasing the positive predictive rate to 0.78 (ref. 9). In doing so, HES permit a cost saving by treating and managing the patients with glaucoma who should be there. Whilst such systems do reduce the number of false positive referrals entering specialist care, the overriding concern is that of scalability. Services are reliant on legacy computer systems where data is transferred by electronic mail, and constrained by the limited clinical time of specialists. Such systems will not be sustainable in years to come given the asynchronous rise in disease prevalence versus the number of available and trained eye care professionals.[10]

The coming of the Artificial Intelligence (AI) age has the potential to transform current models of care, with an emphasis on catalyzing service developments that put patient convenience first. Traditionally, devolving care outside of HES has been associated with a perceived reduction in quality. By utilising classification algorithms, such solutions have the potential to provide quick, accurate decision support at low cost.

AI systems have recently shown great promise in their ability to detect abnormalities from fundus images,[11] breast cancer and melanoma.[12] These systems are all based on so-called deep neural networks. These are loosely inspired by biology; they consist of a number of connected 'neurons', each of which performs simple mathematical operations on its inputs and feeds the results into the connected neurons. Deep learning refers to artificial neural networks where the neurons are arranged in a layered hierarchy; networks with a large number of layers are said to be deep.[13] The layers make it possible to learn a representation of the data with increasing levels of abstraction. For visual tasks, the first layers of the network learn how to represent images as a collection of low-level features such as corners, edges, and blobs. As the levels of abstraction increase, the network is able to represent more complex patterns such as blood vessels or other larger scale features of a fundus image. Deep learning algorithms are able to perform tasks without being explicitly programmed; only the architecture of the neural network is defined, and the system learns how to perform tasks based on the raw image pixels and the corresponding labels (e.g. a diagnosis of glaucoma or normal) for the image. Typically, these systems are trained on thousands of examples.

In a previous study by Seo et al.,[14] the diagnostic performance of the deep learning system (Pegasus v1.0, Visulytix Ltd., London UK) was evaluated and compared to the consensus opinion of two glaucoma specialists. However, the images used were fairly representative of the type of images that the deep learning system was trained on during development, namely they were monoscopic digital fundus photographs with a circular 30 degrees Field of View (FoV). In this study, we further test the generalisation ability of the deep learning system by evaluating it on stereoscopic 15 degrees FoV photographic slides captured in 2003 and scanned into digital format. These types of images were not used in the development of the system. In addition, the performance of the algorithm is compared, retrospectively, to the average performance of a large cohort of ophthalmologists and optometrists from previous studies.[2,15]

The purpose of this study was to evaluate a deep learning system (Pegasus v1.0, Visulytix Ltd., London UK) in the evaluation of stereoscopic optic disc photographs from a previously unseen dataset, and compare its performance to that of ophthalmologists and optometrists measured in prior studies. Deep learning methods for automated diagnosis of glaucoma have previously been developed and validated,[16-18] however the performance of a deployed software product has not previously been evaluated and compared to the performance of a large number of ophthalmologists and optometrists grading the same image data.

# Methods

In this study, the performance of the Artificial Intelligence (AI) system (Pegasus v1.0, Visulytix Ltd., London, United Kingdom) was analysed prospectively, and compared to the average performance of ophthalmologists and optometrists that were determined in the previous studies.[2,15]

## Dataset

The European Optic Disc Assessment Trial (EODAT) was carried out in 2010 and evaluated the assessment of stereoscopic optic disc slides by 243 ophthalmologists across 11 countries.[2] All ophthalmologists were glaucoma specialists. The stereoscopic images were captured by using a 15 degrees Field of View (FoV) Topcon TRC-SS2 (Topcon corporation, Tokyo, Japan). Patients were selected from a cohort of patients and controls who had originally been recruited for a longitudinal glaucoma study at the Rotterdam Eye Hospital. The ophthalmologists were asked to view each stereoscopic slide through a stereo viewer (Asahi Pentax Stereo Viewer II, Hoya Corporation, Tokyo, Japan) and determine a grade; normal or glaucomatous. Importantly, they were given unrestricted time to analyse the images. In 2013, a further study was performed in conjunction with Moorfields Eye Hospital NHS Foundation Trust, which assessed the diagnostic accuracy of 208 optometrists from the United Kingdom.[15]

In both studies, the average diagnostic performance and confidence intervals were determined. In total there were 94 stereo photographic slides; 40 healthy controls, 48 with glaucoma, and 6 with ocular hypertension. The healthy controls and ocular hypertension examples were considered as non-glaucoma for the analysis of Pegasus performance according to the protocols of the previous studies.[2,15]

All photographs of the glaucoma patients were taken from people that were diagnosed with glaucoma by glaucoma specialists based on both the appearance of their optic discs and reproducible visual field defects that were thought to match the appearance of the optic discs. In addition, other clinical information, such as elevated intraocular pressure and family history were taken into account. One eye was randomly selected from each patient if both were eligible for inclusion. The appearance of the optic disc was not used as a selection criterion. No patients from the original studies had significant history of coexisting ocular diseases, histories of intraocular surgery (other than cataract surgery or glaucoma surgery), or systemic diseases that could manifest as visible features in retinal imagery, such as diabetes mellitus. All patients were selected from a cohort of patients that had originally been involved in a longitudinal glaucoma study at the Rotterdam Eye Hospital. All patients were of white ethnic origin.

In the original studies, the patient sample size was chosen such that the number of eyes was large enough to be able to detect differences, but not too large that it would have caused clinicians to lose focus; if the sample size was too large the grading may have been biased from lack of attention. The ophthalmologist and optometrist sample sizes from previous studies were chosen to be large enough to capture the range of diagnostic performances and determine accurate measurements of the average performance of each group. No randomisation was used in the prior studies or the current study, apart from the use of random sampling with replacement for estimating confidence intervals for the performance of the AI system. The investigators were not blinded to the reference standard when performing the analysis of the AI system.

The Rotterdam Eye Hospital institutional human experimentation committee approved the collection and analysis of the data used in this study and previous studies, and informed consent was obtained from the subjects after explanation of the nature and possible consequences of the study.

**AI System Evaluation**

The original stereoscopic slides were scanned into digital format (Fig. 1 and 2), and submitted to the AI system. The scanned images had a resolution of 5,600 x 3,600 pixels, and a file size of approximately 22MB. The diagnostic performance of Pegasus was compared to the performance of ophthalmologists and optometrists determined in the previous studies.

Pegasus is a cloud-deployed clinical decision support software tool for the evaluation of fundus photography and macular Optical Coherence Tomography scans. For fundus image evaluation, the system utilises a collection of CNNs, each specialising in a different task associated with image evaluation, including: key landmark detection (e.g. optic disc, macula and fovea); clinical feature detection (e.g. exudates and haemorrhages); and pathology classification and grading (e.g. diabetic retinopathy). In aggregate, the development dataset consisted of 136,146 anonymised fundus images collected from multiple private sources that suitably reflects variation in observed real-world use cases: patient demographics, types of fundus camera (e.g. manufacturer, FoVs ranging from 30 to 50 degrees, resolution, mydriatic and non-mydriatic), and acquisition artefacts (e.g. dust and scratches). Images used in algorithm development were labelled by between one and five board-certified specialist ophthalmologists from the United Kingdom.

The AI system was not trained on any images captured using a 15 degrees FoV or with the Topcon TRC-SS2. Thus, the dataset used in this study provides a good test of the generalisation ability of the system to unseen image types. The system is designed to generalise to any fundus photograph which contains the optic disc; the system of CNNs first extracts the optic disc from the image, standardises its contrast and size, before feeding it into another CNN that performs a binary classification into *normal* or *abnormal* as well as providing a confidence score. The CNN for abnormality classification is based on the ResNet-50 architecture of He et al.[19] The input size to the classification network is 224-by-224 pixels. During development of the classification CNN, the network was trained from scratch with all parameters randomly initialised at the start of training. Further details of the development of the deep learning system, including architecture and training data used can be found in Supplementary Material S1.

It is not explicitly enforced that the algorithm detects features of the optic disc that are indicative of pathology, instead the algorithm learns to detect important features directly from the raw image pixels after being exposed to a large number of normal and pathological optic discs. It is therefore expected that the CNN "looks" for similar features to clinicians, such as but not limited to: enlargement of the cup; haemorrhages in the direct surrounding of the optic disc; or thinning of the neuroretinal rim. The only information presented to the system is the raw image pixels.

The Pegasus AI system also provides visual explanations that help to contextualise the AI decision making, such as the measurement of the Vertical Cup-to-Disc Ratio (VCDR) and decision heatmaps. The Pegasus decision is intended as an aid to clinicians, enabling them to make faster and more accurate decisions by offering a "second opinion". In this study, the accuracy of the VCDR estimate and the visual explanations were not assessed, only the accuracy of the decision on whether the patient suffers from an optic disc abnormality indicative of glaucomatous optic neuropathy. Pegasus' decision took into account both the confidence that the optic disc contained an abnormality and the estimated VCDR.

No pre-processing was applied to the stereoscopic slides prior to uploading to the Pegasus platform. Unlike the human graders, the AI analysed only one of the images in the stereoscopic pair when making a decision, and did not utilise any stereoscopic information which may help in glaucoma diagnosis. The image "chosen" by the AI was likely to be the image with the best quality view (e.g. best contrast) of the optic disc, since the software extracts the region that it assigns the highest confidence of corresponding to the optic disc prior to disc abnormality and VCDR analysis. The "chosen" image from the pair is made apparent by the visual explanation provided by the Pegasus AI system.

The performance of Pegasus was assessed in terms of the accuracy, specificity, and sensitivity for comparison to the human graders. In addition, the intra-observer agreement for Pegasus was estimated by uploading each image of each stereoscopic pair individually and computing Cohen's Kappa ($\kappa$) coefficient between Pegasus' decision for left and right images of each pair. Since Pegasus is a deterministic system, presenting exactly the same image pixels twice at different times (the protocol used for ophthalmologists and optometrists[2,15]) would yield a perfect intra-observer reliability for the AI system which is uninformative. From Fig. 2, it is evident that there is some difference in appearance between the left and right image of each pair due to, for example, variations in illumination and lens glare artefacts. This variation is similar to re-acquisitions of monoscopic fundus images for the same patient and is therefore suitable for estimating the intra-observer agreement for the AI system.

Confidence intervals, at the 95% significance level, were determined on each of these metrics using 1,000 bootstrap replica samples and the percentile method for confidence interval estimation. Bootstrap methods are convenient since they do not make assumptions on the underlying distribution of the statistic being evaluated. Confidence intervals for human graders have been stated, where possible, based on the results reported in the original studies.[2,15]

# Results

Including the upload time, each image took on average 7 seconds to be analysed by Pegasus. There were no processing failures for any of the images. The authors manually reviewed the images and found that 8 images were of insufficient contrast to make out all of the features of the optic disc, with 3 images being of particularly poor contrast such that the optic disc was difficult to grade. These were included in the performance analysis.

The Pegasus AI system achieved an overall accuracy of 83.7% (95% CI: 77.3-89.1) for the detection of glaucomatous optic neuropathy (Table 1). This was similar to the average accuracy of 243 ophthalmologists and 208 optometrists, as determined in the previous studies.[2,15] Using bootstrap sampling to determine 95% confidence intervals on the difference between Pegasus and the mean accuracy, sensitivity and specificity for optometrists and ophthalmologists, it was determined that there was no statistically significant difference between Pegasus and optometrists nor between Pegasus and ophthalmologists.

Fig. 3 shows the Receiver Operating Characteristic (ROC) curve for Pegasus with the average operating points and confidence intervals plotted for the human graders.

Pegasus shows performance that is within the range of both ophthalmologists and optometrists. The Area Under the Receiver Operating Characteristic curve (AUROC) for the AI system was 87.1% (95% CI: 81.1-92.4).

The intra-observer agreement is given in Table 2, and expressed by using Cohen's Kappa (κ) coefficient. The AI system obtained a Kappa coefficient of 0.74, compared to the values of 0.70 and 0.71 for ophthalmologists and optometrists measured in previous studies, respectively.

# Discussion

Automated analysis of the optic nerve head has significant benefits when embedded within new clinical workflows. The accurate detection of early optic nerve head changes is vital for the detection of early glaucoma. This study is one of the first to evaluate a commercially available AI system to evaluate optic nerve head pathology. Its results suggest an overall accuracy similar to that of the average of European ophthalmologists (glaucoma specialists) and UK optometrists. The relatively high performance of the Pegasus AI system was achieved despite not being able to utilise the stereoscopic information in the photographic slides, unlike the human graders who had access to a stereoscopic viewer. Furthermore, the Pegasus AI was fast at grading images, taking 7 seconds to grade each image (including the time taken to upload the 22MB image files).

Therefore, it is likely that the AI would be faster than humans at analysing the images, which could be helpful in reducing time and expenses in screening programmes.

In addition, the AI system was observed to have good repeatability ($\kappa$=0.74) between the right and left images of each stereoscopic pair, and was similar to that of human graders ($\kappa$≅0.7) from the previous studies.[2,15] In real-life screening environments, the repeatability level for humans is likely to be worse due to fatigue and distraction, and this could contribute to misdiagnosis.

The Pegasus AI system has also shown robustness to a new type of imagery that it had never explicitly seen before (stereoscopic optic disc slides digitised by scanning), demonstrating a good degree of generalisation ability, which has been one of the concerns raised when considering the adoption of AI in healthcare.[20] This is in part possible because the Pegasus AI system has been designed to be robust to a wide range of various fundus imagery by extracting and standardising the optic disc prior to anomaly classification.

A strength of this study was that AI was evaluated on imagery that was previously unseen by the AI system; this type of imagery was not present in either the manufacturer's training, validation or testing of the AI algorithms. In addition, the image quality was typical of clinical practice with approximately 9% of images being difficult to grade due to poor image contrast (as assessed by the authors), making it difficult to observe all of the features in the optic disc in those cases. Another strength of this study was the comparison to an accurate measurement of both ophthalmologist and optometrist performance levels, with a total of 451 human graders having assessed the images for the presence of glaucoma in prior studies.[2,15] Previous studies evaluating the performance of deep learning systems for glaucoma diagnosis have not previously compared the performance of the deep learning systems to both ophthalmologists (glaucoma specialists) and optometrists on the same cohort of patients, and so it has not been clear how well they would fit into the clinical workflow.[16-18] The main limitation of this study was the relatively small sample size, which was, in part, a compromise for the large cohort of human graders. Further studies are required to assess Pegasus performance across a larger number of patients and obtain tighter confidence intervals on the systems performance.

Deep learning based AI systems, such as Pegasus, are potentially transformative - they promise specialist level accuracy for screening and detecting glaucomatous optic neuropathy. They are, in addition, agile in the sense they could be potentially tuned for bespoke performance, for example, to increase specificity to preset values. This would potentially reduce the number of false positives in what is essentially a rare disease. Furthermore, through cloud-based deployment, they can potentially provide such analysis at very low cost to the healthcare provider - irrespective of geographical location. Further investigation, both in terms of clinical and economic analysis, are required as such systems are integrated into real world clinical practice.

# Acknowledgements

The authors acknowledge Visulytix Ltd. for supply of hardware and funding.

# Conflicts of Interest

# Funding

Funding for the analysis of the EODAT images was provided by Visulytix Ltd.

# Figures and Tables

**Table 1** Diagnostic performance of ophthalmologists, optometrists and Pegasus AI. Confidence intervals were determined at the 95% significance level. Performance measures quoted for ophthalmologists and optometrists are the mean and median performance, respectively.[2,15]

|  | Accuracy (%) | Specificity (%) | Sensitivity (%) |
|---|---|---|---|
| Mean ophthalmologist[2] | 80.5 (95% CI: 67.2-93.8) | 87.4 (95% CI: 65.8-100) | 74.7 (95% CI: 53.9-95.5) |
| Median optometrist[15] | 80 (95% CI: 67-88) | 74 (95% CI: 62-88) | 92 (95% CI: 70-100) |
| Pegasus AI* | 83.7 (95% CI: 77.3-89.1) | 86.2 (95% CI: 78.3-93.2) | 80.9 (95% CI: 71.4-89.6) |

* Confidence intervals were estimated by bootstrap sampling with 1,000 replicas.

**Table 2** Intra-observer agreement. Agreement measures quoted for ophthalmologists and optometrists are the mean and median agreement, respectively. The intra-observer agreement for Pegasus was determined by analysing the left and right image of each stereoscopic pair separately for all pairs and using bootstrap sampling with 1,000 replicas to determine 95% confidence intervals.

|  | Cohen's Kappa, κ (range) |
|---|---|
| Ophthalmologists | 0.70<br>(range: -0.13-1.00; 95% CI: 0.67-0.73) |
| Optometrists | 0.71<br>(range: 0.08-1.00) |
| Pegasus AI | 0.74<br>(95% CI: 0.63-0.85) |

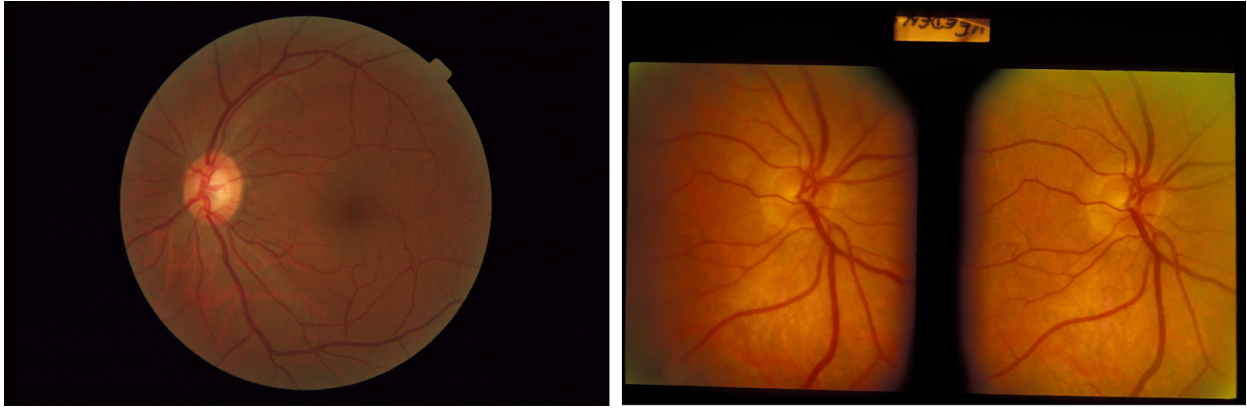

**Fig. 1** Conventional 45 degrees Field of View (FoV) fundus image typical of the dataset used to train Pegasus (left), and a scanned stereoscopic optic disc slide from the EODAT.

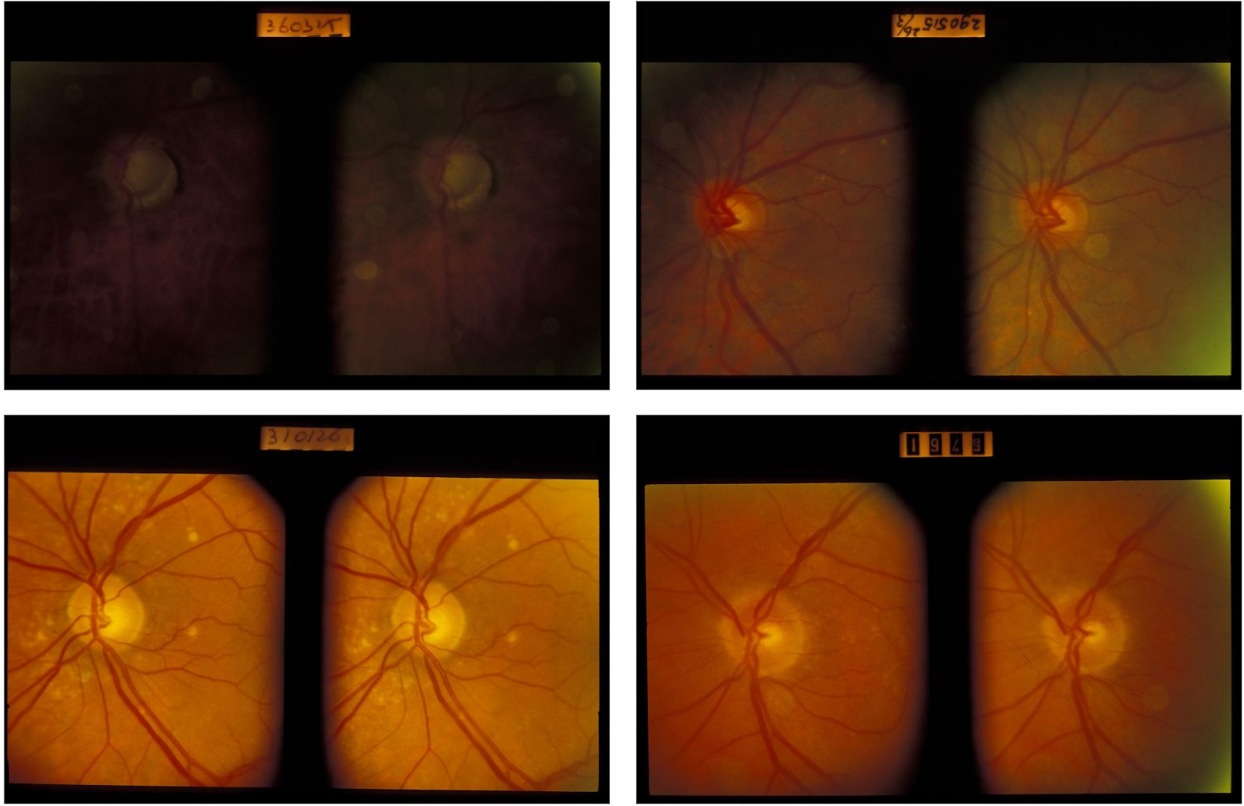

**Fig. 2** Representative examples of images from the European Optic Disc Assessment Trial (EODAT)[2] that were used in this evaluation.

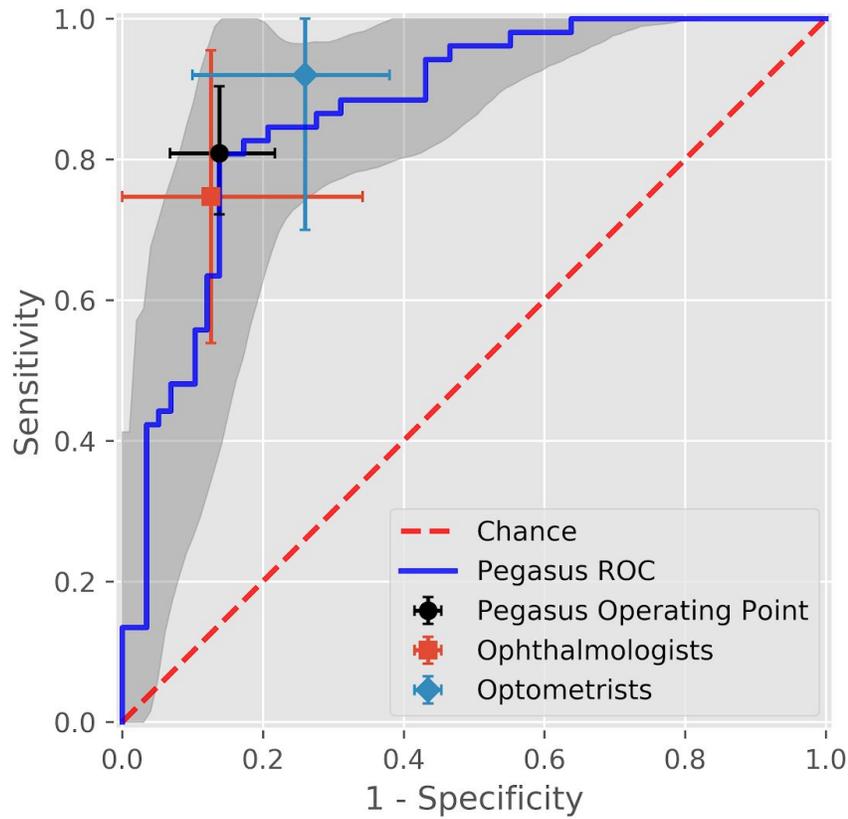

**Fig. 3** Receiver Operating Characteristic (ROC) curve for Pegasus. The shaded region indicates the 95% confidence interval on the Pegasus ROC estimated using bootstrap sampling with 1,000 replications. The average performance and 95% confidence intervals for ophthalmologists (square), optometrists (diamond), and the Pegasus operating point (circle) are plotted.